\pdfoutput=1  

\documentclass[11pt,a4paper]{article}
\usepackage[T1]{fontenc}
\usepackage[utf8]{inputenc}
\usepackage{times}
\usepackage{latexsym}
\usepackage{microtype}
\usepackage{graphicx}
\usepackage{booktabs}
\usepackage{amsmath}
\usepackage{xcolor}
\usepackage{natbib}
\usepackage[breaklinks=true,colorlinks=true,linkcolor=blue,citecolor=blue,urlcolor=blue]{hyperref}
\usepackage{url}
\usepackage{xurl}  
\usepackage[margin=2.5cm]{geometry}
\usepackage[section]{placeins}  

\graphicspath{{figures/}}

\title{HistoRAG: Embedding Historical Methodology in\\ Retrieval-Augmented Generation\\ Through Critical Technical Practice}

\author{
  Noah J. Kim-Baumann \quad Torsten Hiltmann \\
  Humboldt-Universit\"at zu Berlin \\
  \texttt{digitalhistory@hu-berlin.de}
}

\date{}

\begin{document}
\maketitle

\begin{abstract}
Retrieval-Augmented Generation (RAG) is the prevailing architecture for grounding language model outputs in external evidence, yet its dominant evaluation paradigms and default configurations remain oriented toward factual question-answering. For interpretive disciplines such as historical studies, RAG embeds assumptions that conflict with scholarly practice. We thus introduce HistoRAG, a framework that translates historiographical principles into concrete architectural interventions. \textit{Separated retrieval and generation} decouples source discovery from interpretation, \textit{temporal windowing} enforces balanced source representation across the research period as a methodological requirement of historical inquiry, and \textit{LLM-as-judge evaluation} makes relevance judgments transparent and contestable. We evaluate these interventions using SPIEGELragged, an implementation applied to 102,189 articles from the news journal \textit{Der Spiegel} (1950--1979). Our evaluation shows that each intervention addresses a measurable deficiency in standard RAG: era-specific vocabulary retrieves zero chunks from the 1950s when using 1970s terminology, evidence of the temporal skew that motivates windowing; vector similarity and LLM-assessed relevance correlate only weakly (Spearman $\rho$ = 0.275), motivating post-retrieval evaluation; and keyword-based and semantic retrieval surface largely disjoint source pools, motivating an architecture in which both operate as complementary retrieval layers under a shared LLM evaluation filter. Additionally, we introduce the concept of \textit{Zwischentexte} (intermediate texts that function as interpretive proposals rather than findings) as a framework for responsible integration of LLM-generated text into scholarly practice. The architecture we propose offers a model for how domain-specific epistemological commitments can be translated into RAG design decisions, with potential to transfer to other interpretive disciplines working with large corpora.
\end{abstract}

\section{Introduction}

Retrieval-Augmented Generation has become an integral piece of LLM-application architectures, used for grounding language model outputs in external evidence. Yet RAG system design often remains generic, oriented toward factual question-answering, optimised to find the single best passage, generate a concise answer, and move on. By design these systems tend to encourage frictionless user behaviours no matter the source type. For disciplines whose engagement with text is fundamentally interpretive, standard RAG architectures embed assumptions that conflict with hermeneutic scholarly practice. In such fields, "relevance" is empirically hard to define, often depending on the individual researcher's specific epistemic interest, the research questions being pursued, temporality, and semantics, and the basis for selecting sources must itself be open to scrutiny. The task here is not information retrieval in the conventional sense of locating statements within a corpus, but interpretation. It means posing questions about the texts themselves, such as how the statements they contain shift over time, which amounts to making statements about those statements rather than merely finding them. This paper asks how RAG can be redesigned to support historical research, and demonstrates that doing so requires architectural decisions grounded in the epistemological commitments of the discipline.

Digital History is currently undergoing a transition that is both fundamental and disruptive. For decades, computational methods have offered historians powerful tools for analysing large corpora, with methods that, as we argue, generally share a focus of operating on sequences of characters, at the level of signs rather than meaning. Drawing on Silke Schwandt's characterisation of the computer as "semantically blind" \citep{schwandt2018DigitaleMethoden}, these earlier computational methods operated on the surface of the text rather than its meaning. As we have argued \citep{hiltmann2021DigitalMethodsPractice}, the deeper point is that the semantic level was not available to these methods at all, a constraint that shaped what historians could do with them. We characterise this paradigm as \textit{DH~1.0}. For example, topic modelling operates on co-occurrence matrices, text reuse detection identifies repeated character sequences, and corpus tools like Voyant \citep{rockwell2016HermeneuticaComputerassistedInterpretation} visualise frequencies of token sequences. Large Language Models (LLMs) now promise something qualitatively different, a shift toward what we call \textit{DH~2.0}. LLMs work with what \citet{simons2025LargeLanguageModels} call "contextualised representations", which are computational approximations of meaning that enable both pattern recognition and interpretive generation (p.\ 2). While LLMs still operate on tokens internally, the crucial difference is that our interaction with these tools occurs at the level of meaning rather than formal operations. The historian poses questions in natural language and receives responses that engage with semantic content, not frequency tables or co-occurrence matrices. This shift from computing sequences of characters to computing approximations of the semantic meaning of those sequences raises urgent questions for interpretive disciplines, and represents a fundamental epistemological change.

This potential comes with real dangers. If earlier methods required historians to explicitly formalise their analytical approaches and interpret the results, LLMs risk hiding these choices entirely. When a system can "read" thousands of sources and generate coherent analyses, how do we preserve the interpretive authority that defines historical scholarship? As \citet{chen2025ToolsGenerativeAI} argues, this is at its core a question of epistemic agency. It is a matter of preserving researchers' capacity to make, evaluate, and contest knowledge claims instead of delegating these to computational systems. The risk is not that AI will replace historians but that we will accept interfaces that obscure the critical decisions that constitute scholarly practice (such as source selection, relevance evaluation, interpretive framing). If we do not actively shape the tools we use, commercial vendors will design them for us, optimising for efficiency and seamless user experience at the expense of epistemic transparency.

RAG offers a path forward because it preserves the distinction between the reasoner (the LLM) and the archive (the documents), allowing generated insights to be traced back to specific sources \citep{lewis2020RetrievalAugmentedGenerationKnowledgeIntensive}. Unlike fine-tuning, which buries knowledge within neural network weights and produces outputs whose epistemic origins cannot be traced, RAG maintains the connection between claims and evidence that historical scholarship demands. As \citet{weatherby2025LanguageMachinesCultural} argues, LLM outputs are cultural productions shaped by training data in ways that resist inspection. RAG offers a means of preserving the link between generated text and identifiable sources. But standard RAG architectures are typically built and evaluated for information retrieval rather than interpretive inquiry. They conflate source discovery with interpretation in a seamless pipeline, treat similarity as a proxy for relevance, lack mechanisms for ensuring temporal coverage, and treat the vocabulary of the corpus as stable, ignoring the conceptual change that historical inquiry must track. These are all assumptions that conflict with the requirements of historical method.

We introduce HistoRAG,\footnote{\url{https://scm.cms.hu-berlin.de/digital-history/digital-history-forschung/histo_rag/historag}} a framework that embeds historical methodology into RAG architecture through three interventions grounded in Critical Technical Practice \citep{agre1998CriticalTechnicalPractice}. First, \textbf{separated retrieval and generation} formally decouples corpus construction from analysis, restoring the distinction between \textit{Heuristik} (source discovery) and \textit{Interpretation} that historical method demands. Second, \textbf{temporal windowing} enforces balanced representation of sources across the research period as a methodological requirement of historical inquiry, guaranteeing a quota of sources per time window rather than allowing similarity ranking to determine temporal coverage. Standard retrieval introduces temporal skew through two mechanisms: vocabulary drifts across decades so contemporary query terms align more closely with later periods, and thematic density varies over time so periods with concentrated discussion of a topic crowd out comparably relevant sources from sparser periods. Third, \textbf{LLM-as-judge evaluation} introduces historian-defined criteria into post-retrieval filtering, turning algorithmic selection from a black box into a transparent, argumentative process whose justifications can be reviewed and contested. We evaluate these interventions using SPIEGELragged,\footnote{\url{https://scm.cms.hu-berlin.de/digital-history/digital-history-forschung/histo_rag/spiegelragged}} an implementation applied to 102,189 articles from \textit{Der Spiegel} (1950--1979), demonstrating that each intervention addresses a measurable deficiency in standard RAG, and that the keyword filtering and semantic retrieval layers the architecture also incorporates capture largely complementary dimensions of relevance under a shared LLM evaluation filter.

While our case study is historical, the underlying problem extends to other interpretive disciplines working with large corpora, including literary studies, legal analysis, qualitative social science, archival research, and others. Wherever "relevance" is not a fixed property of documents but a function of the question being asked, wherever temporal or contextual positioning shapes meaning, and wherever the basis for source selection must be transparent and contestable, the tension between standard RAG design and disciplinary practice recurs. With HistoRAG, we offer a model for how domain-specific epistemological commitments can be translated into concrete architectural decisions.

\section{Related Work}

HistoRAG sits at the intersection of three research trajectories. The first is the evolution of RAG architectures beyond factual question-answering. The second is the growing recognition that domain-specific corpora demand domain-specific retrieval designs. The third involves the methodological debates within digital humanities about how computational systems intervene in interpretive scholarship. We trace these trajectories here to identify the specific gap our framework addresses.

Retrieval-Augmented Generation, introduced by \citet{lewis2020RetrievalAugmentedGenerationKnowledgeIntensive}, was designed to ground language model outputs in external evidence, combining parametric memory with a searchable knowledge base. Subsequent work has refined this architecture considerably. \citet{gao2024RetrievalAugmentedGenerationLarge} distinguish between "naive" and "advanced" RAG pipelines, identifying post-retrieval processing (including re-ranking and filtering) as a key site of improvement. \citet{huang2024SurveyRetrievalAugmentedText} organise the field into four phases (pre-retrieval, retrieval, post-retrieval, generation), noting that most enhancements target retrieval accuracy for factual lookup tasks. The "LLM-as-a-judge" paradigm, in which instruction-following models evaluate content against explicit criteria, has emerged as a promising post-retrieval strategy \citep{gu2024SurveyLLMasaJudge}, and \citet{nogueira2020PassageRerankingBERT} demonstrated that BERT-based passage re-ranking substantially outperforms initial retrieval on standard benchmarks. Yet across this literature, the dominant evaluation paradigm remains oriented toward information retrieval benchmarks, where systems are assessed on whether they return the single best passage to answer a factual question. \citet{murugadoss2024EvaluatingEvaluatorMeasuring} found that even highly specific scoring instructions improve LLM-human alignment by only $\sim$4\% (p.\ 2), and that model perplexity sometimes aligns better with human judgments than prompting does, underscoring the limits of prompt-based evaluation for factual tasks.

In the years since Lewis et al.'s original proposal, a growing body of work has demonstrated that generic RAG pipelines falter when applied to specialised domains. In medicine, \citet{kim2025RethinkingRAGMedicine} conducted the largest expert evaluation of medical RAG to date, finding that standard RAG often degraded rather than improved performance, with only 22\% of retrieved passages rated relevant by expert evaluators. A systematic review of 70 healthcare RAG studies similarly concluded that basic implementations yield only marginal improvements and that effective clinical applications require domain-adapted embeddings, structured knowledge integration, and expert validation \citep{amugongo2025RAGHealthcareSystematic}. In law, \citet{magesh2025HallucinationFreeAssessingReliability} demonstrated that commercial legal RAG tools hallucinate between 17\% and 33\% of the time, indicating that domain-focused applications require not just improved retrieval but perhaps different frameworks entirely. Additionally, these efforts show that the distance between general-purpose RAG and disciplinary requirements is not merely quantitative (more documents, better embeddings) but qualitative, involving different conceptions of what "relevance" means in context.

Within digital humanities and historical research specifically, engagement with RAG remains nascent but is accelerating. \citet{murugaraj2025TopicRAGHistoricalNewspapers} introduced TopicRAG, integrating BERTopic-based topic modelling with retrieval to improve thematic coherence when searching historical newspaper archives. Their system restricts retrieval to topically relevant documents, accepting reduced recall as a trade-off for precision. This strategy contrasts with our own, which treats \textit{recall for precision} as a central design principle. Because relevance in historical inquiry is interpretive and contestable, we do not let a similarity threshold discard sources before a historian can judge them. HistoRAG instead casts a wide net at retrieval and achieves precision afterwards, through a transparent evaluation step that keeps the relevance decision with the researcher. \citet{lee2025RetrievalAugmentedGenerationSystem} applied RAG to the Annals of the Joseon Dynasty, a 472-year historical record, demonstrating substantial improvements over LLM-only baselines while emphasising the need for clear source citation as a non-negotiable requirement of historical analysis. Most directly relevant to our work, \citet{zhou2025HumanitiesintheLoopUsingClose} proposed a "humanities-in-the-loop" framework that embeds close reading practices into RAG pipeline stages for digital archival materials. Their argument that conventional RAG fails to capture the contextual complexity and interpretive nuance required by humanities research closely parallels our own, though their focus on close reading as method differs from our emphasis on integrating historiographical theory (periodisation, source criticism, temporal reasoning) into system architecture. Beyond RAG specifically, \citet{brausch2023MachineLearningHistory} have explored how machine learning can combine computational breadth with traditional depth in the history of ideas, while broader critical assessments of AI in archival science \citep{shinde2025TracingPastPredicting} emphasise the need for interdisciplinary collaboration between computational and domain experts.

These emerging applications of RAG and LLMs to historical materials share the recognition that generic architectures are insufficient, but they have not yet systematically addressed the \textit{epistemological} dimensions of the problem. This is where a separate methodological literature becomes essential. A first strand argues that computational systems must be shaped from within their disciplines of application rather than merely critiqued from outside. \citeauthor{agre1998CriticalTechnicalPractice}'s \citeyearpar{agre1998CriticalTechnicalPractice} concept of Critical Technical Practice provides the foundational design philosophy, recently extended by \citet{hirsbrunner2024CriticalTechnicalPractice} into "Reflexive Data Science". \citet{fickers2022DigitalHistoryHermeneutics} develop a complementary programme of "digital hermeneutics", arguing that computational infrastructures intervene in research practices at every phase (from search through interpretation to publication) and that these interventions must be made visible, not hidden behind seamless interfaces (p.\ 8). A second strand addresses what happens when LLMs specifically enter this landscape. \citet{hiltmann2021DigitalMethodsPractice}, adopting \citeauthor{schwandt2018DigitaleMethoden}'s \citeyearpar{schwandt2018DigitaleMethoden} description of the computer as "semantically blind", characterise earlier computational methods as operating in the "realm of signal" rather than the "realm of meaning", a characterisation that LLMs partially challenge but do not resolve, since their semantic operations remain opaque in ways that sequence-based methods were not. \citet{chen2025ToolsGenerativeAI} frames this opacity as a question of epistemic agency, while \citet{simons2025LargeLanguageModels} identify a related "accessibility-literacy trade-off" in which natural language interfaces lower barriers to entry while obscuring the interpretive assumptions embedded in retrieval algorithms and prompt templates. Finally, \citeauthor{welskopp2008HistorischeErkenntnis}'s \citeyearpar{welskopp2008HistorischeErkenntnis} foundational account of historical method, particularly his concept of the \textit{Kontinuit\"atsannahme} (the principle that historical understanding requires tracing connections across time), provides the specific theoretical grounding for our temporal windowing intervention.

What remains absent from this landscape is a framework that systematically translates historiographical principles into RAG architecture decisions. Domain-specific RAG systems in medicine and law have adapted retrieval and evaluation to disciplinary needs, though they do not typically frame these adaptations in terms of explicit epistemological theory in the way humanities scholarship demands. Historical RAG applications have demonstrated the potential of grounding LLM outputs in archival sources, but have not yet redesigned the pipeline around the specific requirements of historical method, including temporal reasoning, source criticism, and the interpretive evaluation of relevance. And the methodological literature on digital hermeneutics and critical technical practice has articulated what responsible AI-supported scholarship could look like, without offering concrete architectural implementations. HistoRAG aims to bridge these three strands by translating the epistemological commitments identified by Agre, Fickers, Welskopp, and others into specific design interventions that reshape how a RAG system operates when supporting historical research.

\section{The HistoRAG Framework}

HistoRAG transforms RAG from an information retrieval system into a framework for historiographical inquiry. Rather than optimising retrieval accuracy for factual question-answering, we redesign the pipeline around principles drawn from historical methodology. The interventions described below emerged from an iterative process of building, testing, and refining the system against actual historical research questions. Each addresses a specific point where standard RAG architectures embed assumptions that conflict with the requirements of interpretive scholarship, and each is grounded in a distinct historiographical commitment elaborated in Section~\ref{sec:design_principles}.

\subsection{Design Principles}
\label{sec:design_principles}

Our design philosophy follows \citeauthor{agre1998CriticalTechnicalPractice}'s \citeyearpar{agre1998CriticalTechnicalPractice} concept of Critical Technical Practice. For RAG and historical research, this means that the discipline must propose its own architectures rather than accepting designs optimised for other purposes. As \citeauthor{fickers2022DigitalHistoryHermeneutics}'s \citeyearpar{fickers2022DigitalHistoryHermeneutics} programme of "digital hermeneutics" makes clear, every technical decision in a computational research infrastructure is also an epistemological one. Applied to RAG, this means recognising that architecture \textit{is} epistemology. How we chunk texts, rank results, select models, and formulate prompts are all decisions that shape what knowledge can be produced. Making these decisions explicit and contestable is itself a core design requirement.

Each intervention is grounded in a specific historiographical commitment. First, historical method distinguishes between \textit{Heuristik} (the systematic discovery and gathering of sources, together with the sharpening of the research question that guides them) and \textit{Interpretation} (the construction of meaning from assembled evidence). This distinction was formalised in Droysen's \textit{Historik} (1857; see \citealp{maclean1982JohannGustavDroysen}) and has been maintained across historiographical traditions. Standard RAG collapses it into a single pipeline. Second, \citeauthor{welskopp2008HistorischeErkenntnis}'s \citeyearpar{welskopp2008HistorischeErkenntnis} \textit{Kontinuit\"atsannahme} (continuity assumption) holds that historical understanding requires tracing connections across time; this temporal commitment has no equivalent in similarity-based retrieval. Third, historical claims must remain connected to their evidentiary foundations (what German historiography calls \textit{Quellenbeleg}, or source citation) and the basis for selecting and evaluating historical source material must itself be open to scrutiny. These principles translate into the concrete architectural decisions described below.

\begin{figure}[htbp]
  \centering
  \includegraphics[width=0.95\textwidth]{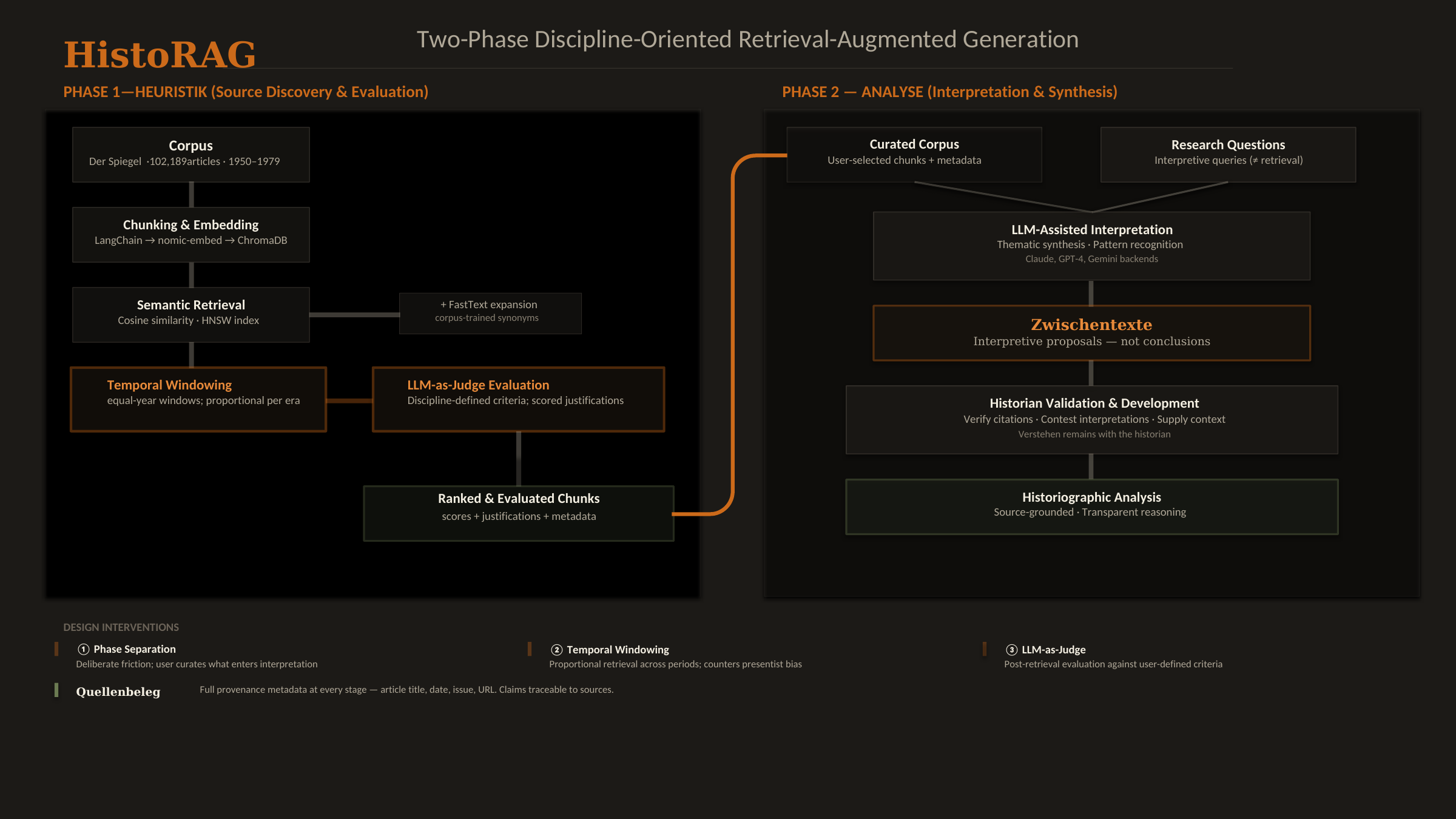}
  \caption{The HistoRAG two-phase pipeline. The \textit{Heuristik} phase (left) handles source discovery through temporal windowing and LLM-as-judge evaluation; the \textit{Analyse} phase (right) enables LLM-assisted interpretation of curated sources. Explicit user-initiated transfer between phases preserves the historian's control over which sources enter interpretation.}
  \label{fig:architecture}
\end{figure}

\subsection{Separated Retrieval and Generation}
\label{sec:separated_retrieval}

Standard RAG systems flow seamlessly from query to retrieval to generation. A user's question triggers similarity search, retrieved chunks feed directly into an LLM, and a response appears. This is optimised for speed and convenience but conflates two distinct scholarly activities, namely finding relevant sources and asking questions of those sources.

HistoRAG formally separates these phases into \textit{Heuristik} (retrieval and corpus construction) and \textit{Analyse} (LLM-assisted analysis of curated sources), as shown in Figure~\ref{fig:architecture}. Transfer between phases requires an explicit user action, preventing the seamless automation that would bypass the historian's judgment about which sources merit interpretive engagement.

The critical design insight underlying this separation is that the \textbf{retrieval query} and the \textbf{analysis question} serve fundamentally different purposes. The retrieval query aims to cast a broad net across the corpus, and it takes the form of a topic description rather than a question, since its purpose is to delimit a region of the corpus to gather rather than to be answered. Descriptive phrasing works well here because our system attaches the embedding model's query prefix automatically, so the input is treated as a search query and matched against documents whether or not it is phrased interrogatively \citep{nussbaum2025NomicEmbedTraining}. For example, "West German media coverage of automation, computers, and technological unemployment 1950--1970" describes the region of the corpus to be searched. The analysis question, by contrast, guides interpretation of assembled sources. It can be specific, comparative, or exploratory in ways that would be counterproductive for retrieval, such as "How did the assessment of automation and computerisation change over the period?" or "Identify shifts in the vocabulary used to describe computers across these sources."

This mirrors traditional archival practice. A historian searching a catalogue for "Mauer Berlin 1961" will later ask very different questions of the assembled sources, questions about representation, rhetoric, and discursive evolution that operate in a different register from the search terms. Standard RAG collapses these registers into a single query, either narrowing retrieval (missing sources that don't match the interpretive framing) or blurring analysis (providing no guidance for how to read the assembled sources). Our separation preserves the distinction computationally. An agentic system could, of course, automatically decompose a user's interpretive question into one or more retrieval queries, and such decomposition is compatible with our architecture. But we argue that making the separation explicit has independent value: it prompts the researcher to reflect on what they are searching for versus what they want to know, a distinction that automated decomposition would render invisible (see Section~\ref{sec:agentic} for a broader discussion).

Between the two phases, the historian reviews retrieved chunks, examines metadata, and curates a working corpus by accepting, rejecting, or flagging sources before any LLM-assisted interpretation begins. This manual curation step positions algorithmic retrieval as a \textit{first filter} that surfaces candidates for human review, preserving interpretive authority while enabling computational reach. The deliberate friction is a feature, not a limitation. It ensures that the transition from source discovery to interpretation remains a conscious methodological decision, not an automated pipeline step.

\subsection{Temporal Windowing}
\label{sec:temporal_windowing}

Historical research on questions of change across time requires a source sample that is balanced across the period under investigation. This is what historians formalise as \textit{Stichproben}: deliberate, stratified sampling across the full temporal span rather than reliance on whatever density of sources happens to surface for a given query. \citeauthor{welskopp2008HistorischeErkenntnis}'s \citeyearpar{welskopp2008HistorischeErkenntnis} \textit{Kontinuit\"atsannahme} expresses the underlying commitment: historical understanding requires tracing connections across time, which presupposes that sources from across that time are actually in front of the historian to begin with. Similarity-based retrieval, optimised for the most proximate passages rather than for temporal coverage, has no such commitment, and in a corpus spanning decades it systematically produces temporally skewed source sets through two distinct mechanisms.

The first is \textit{vocabulary drift}. Discourse evolves over time \citep{hamilton2016DiachronicWordEmbeddings, kutuzov2018DiachronicWordEmbeddings}, and chunks from periods where terminology most closely matches contemporary query formulations will be systematically over-represented, while earlier periods (where the same phenomena were described in now-unfamiliar language) will be under-retrieved. The second is \textit{thematic density}. Topics are not discussed at constant intensity across decades; in periods where a topic is concentrated, many highly similar chunks cluster at the top of the ranked retrieval list, pushing comparably relevant but less densely surrounded chunks from sparser periods further down and, in practice, off the result set entirely.

Neither mechanism is a flaw in the embedding model; both follow from a faithful capture of semantic proximity. The problem is epistemological. For questions about discourse evolution, high semantic similarity to a modern query may inversely correlate with historical significance. The earliest sources, precisely those that document the emergence and transformation of concepts, use vocabulary most distant from contemporary usage; conversely, the periods most densely covered by the press are often not the periods in which the formative ideas were first articulated. Standard RAG thus embeds what we call a \textit{vocabulary alignment bias} and, alongside it, a \textit{thematic density bias}, both systematic tendencies to let whatever sources happen to dominate the similarity ranking displace the temporally balanced sample that historical inquiry requires.

Temporal windowing addresses both biases through a simple but consequential design choice. Rather than retrieving the top-$N$ most similar chunks across the entire corpus, we divide the search period into temporal windows (e.g., five-year intervals) and retrieve equally from each. For a thirty-year corpus with six windows and twenty chunks per window, the system guarantees balanced temporal representation regardless of how vocabulary similarity or thematic density distributes across time.

This intervention makes computationally explicit a stratified-sampling logic that historians otherwise apply intuitively. Importantly, windowing operates on a different dimension from query expansion, and HistoRAG uses both as complementary measures against the temporal skew that similarity retrieval introduces. Query expansion, which our architecture incorporates via corpus-trained word embeddings (see Section~\ref{sec:system_architecture}), enlarges the lexical reach of a single search to surface synonyms and period-specific variants of the query terms. Windowing redistributes the temporal composition of what the search returns. The two address overlapping but distinct facets of the same underlying problem. Standard RAG systems incorporate neither, and commercial implementations offer no mechanism for the temporal-balance dimension. As \citet{welskopp2008HistorischeErkenntnis} argues, ``die Methode folgt aus der Frage und der Theorie'' (method follows from question and theory). There is no universal optimum, only choices appropriate to specific inquiries (p.\ 132). The trade-off is explicit: some windows will yield lower-similarity chunks than a global search would return, and for temporal windows covering early periods, we retrieve chunks at lower similarity scores while potentially excluding higher-scoring chunks from later periods. This is a methodological choice, not an error, and it is visible and contestable. It is also our \textit{recall for precision} principle in operation. By keeping lower-similarity sources from every window rather than letting a global ranking discard them, the system preserves material that a precision-first search would lose, so the historian can examine it and build the set most relevant to their specific research question, drawn from across the full period and including sources that bear on it only indirectly. For historical interpretation, missing a relevant source is far more costly than reviewing one that turns out to be marginal. The historian can adjust window sizes, modify retrieval counts per window, or disable windowing entirely for questions where temporal balance is not relevant.

\subsection{LLM-as-Judge: Post-Retrieval Evaluation}
\label{sec:llm_judge}

Separated retrieval ensures the historian controls which sources enter interpretation; temporal windowing ensures those sources are balanced across the research period, with neither vocabulary alignment nor thematic density allowed to determine temporal coverage. But neither addresses a more fundamental problem: whether the retrieved sources are actually \textit{relevant} to the research question being asked. Semantic similarity, even within temporal windows, remains a blunt instrument for assessing interpretive relevance. Vector similarity captures lexical and conceptual proximity but cannot evaluate whether a source addresses the specific dimensions that matter for a given research question. A journalistic report \textit{about} public fears of automation is not the same as a reader letter that itself \textit{articulates} fear. A query embedding has limited capacity to encode such nuanced criteria; adding affective terms like "Hoffnung" (hope) or "Angst" (fear) alongside "Computer" and "Automatisierung" dilutes the semantic focus rather than sharpening it.

Our third intervention introduces an LLM-based evaluation layer after retrieval and temporal balancing. A language model reads each retrieved chunk and evaluates it against explicit, researcher-defined criteria, producing both a numerical score and a written justification citing specific textual evidence. This post-retrieval step draws on the "LLM-as-a-judge" paradigm \citep{gu2024SurveyLLMasaJudge} but adapts it for interpretive rather than factual evaluation. Research on LLM-based evaluation has shown that even highly specific scoring instructions yield only marginal improvements in LLM-human alignment for factual tasks \citep{murugadoss2024EvaluatingEvaluatorMeasuring}. We draw a different implication for interpretive disciplines: where "relevance" is itself a contested category, the value of explicit evaluation criteria may lie less in improving accuracy than in making the \textit{basis} of relevance judgments transparent and open to scholarly contestation.

The evaluation criteria are designed by the historian, not the system. For a research question such as public attitudes toward computerisation, the criteria specify that highest scores go to sources that explicitly thematise hopes or fears, contain voices of affected individuals, or represent public debate. These are the interpretive qualities that distinguish sources valuable for understanding attitudes from sources that merely mention technology. The criteria are formulated in natural language as a detailed rubric (scored 1--10), making the basis for relevance judgments transparent and open to scholarly contestation. A different researcher with different theoretical commitments could substitute different criteria, producing different evaluations of the same corpus, which is precisely how interpretive scholarship should work.

Three prompt-design decisions merit attention. First, the system requires \textit{argumentation before scoring}. The LLM must justify its assessment with reference to specific textual evidence before assigning a numerical score. This produces evaluation texts that function as interpretive annotations, not opaque relevance labels. The historian can examine \textit{why} a source was rated highly or poorly, enabling informed acceptance or rejection of the algorithmic judgment. Second, we enforce \textit{absolute rather than comparative evaluation}. During development, we found that batch processing, in which several sources are evaluated together in a single request to the model, led models to rank sources relative to each other within a batch instead of evaluating each against the stated criteria. This is consistent with documented biases in LLM-based evaluation, where comparative settings introduce position and preference effects that independent, criterion-referenced scoring avoids \citep{gu2024SurveyLLMasaJudge}. The evaluation prompt explicitly instructs independent assessment, producing more consistent scores across different batch configurations. Third, the criteria foreground dimensions that vector similarity cannot capture, such as explicit thematisation, affective framing, and voices of affected individuals. These encode the historian's scholarly interest (in our case, in vernacular discourse and reader responses) as a computational filter.

The result is a three-layer retrieval architecture in which keywords, vector similarity, and LLM interpretation function as complementary filters, not competing alternatives. Keywords filter by vocabulary, vector similarity filters by conceptual proximity, and LLM evaluation filters by interpretive relevance to the specific research question. As we demonstrate in Section~\ref{sec:evaluation}, this multi-layer approach meaningfully changes which sources a historian engages. In our evaluation, keyword filtering, semantic retrieval, and LLM evaluation capture largely complementary dimensions of relevance, with the LLM evaluation surfacing interpretively valuable sources (particularly vernacular texts like reader letters) that neither keyword search nor similarity retrieval alone would prioritise.

\section{Implementation: SPIEGELragged}

We implement HistoRAG as SPIEGELragged, a web-based research application for investigating discourse patterns in \textit{Der Spiegel}, West Germany's leading news weekly, across the period 1950--1979. The system, corpus, and research question together constitute the testbed for evaluating the framework's interventions. 

The framework and the application are released as two separate codebases (see Code and Data Availability). The framework, distributed as the Python package \verb|historag|, contains the reusable components: search strategies, multi-provider LLM dispatch, the streaming API, the \verb|Corpus| adapter protocol, and Docker deployment templates. SPIEGELragged, on the other hand, is a thin consuming layer on top of the framework. It supplies a SPIEGEL-specific \verb|Corpus| subclass that implements the framework's adapter protocol, along with German historian system prompts, the \textit{Der Spiegel} metadata schema, ChromaDB collection naming, ingest scripts, and the user-facing frontend. This separation is itself an instance of the design principle the paper advocates. It keeps the architecture inspectable and lets other research projects build on the framework by implementing their own \verb|Corpus| adapter, as SPIEGELragged does for \textit{Der Spiegel}, rather than forking it. The framework thus follows a plugin (microkernel) architecture, a reusable core that exposes a single well-defined extension point. This is a direct application of the dependency inversion principle, in which the core and each application both depend on the \verb|Corpus| abstraction rather than on one another \citep{martin2003AgileSoftwareDevelopment}, and it reflects the FAIR principles for research software, under which software should be reusable, able to be understood, built upon, or incorporated into other software \citep{barker2022FAIRPrinciplesResearchSoftware}.

\subsection{Corpus and Research Context}
\label{sec:corpus}

\textit{Der Spiegel} offers a particularly productive corpus for testing HistoRAG's design. Founded in 1947, the magazine reached approximately 490,000 copies in circulation by 1961 \citep{enzensberger1962Einzelheiten1BewusstseinsIndustrie}, with an estimated readership of two million, drawn predominantly from what Enzensberger (p.\ 63) termed "meinungsbildende Gruppen" (opinion-shaping groups including teachers, journalists, politicians, and professionals). Our digitised archive, scraped from the \textit{Der Spiegel} online archive, comprises 102,189 articles across thirty years, a scale that exceeds what any individual researcher could systematically engage through close reading while remaining tractable for computational analysis.

The corpus's temporal span (1950--1979) captures computerisation's trajectory from early mainframes through to the threshold of personal computing \citep{schmitt2016DigitalgeschichteDeutschlandsForschungsbericht}. Crucially for testing temporal windowing, vocabulary shifted substantially across this period. Early coverage used anthropomorphising terms like "Elektronenhirn" (electronic brain), which gave way to "Computer" by the late 1960s and bureaucratic abstractions like "EDV" (elektronische Datenverarbeitung, electronic data processing) in the 1970s \citep{busch2015DiskurslexikologieUndSprachgeschichte}. This vocabulary drift creates exactly the retrieval bias that temporal windowing is designed to address.

We operationalise our research through two metaquestions. The first tracks terminology evolution. How did the language used to describe computing shift across three decades, and did different article types (editorial coverage, reader letters) adopt new terminology at different rates? The second examines public sentiment. How did hopes and fears associated with computerisation and automation evolve, and did reader letters reflect, anticipate, or resist editorial framing? Both questions demand comprehensive temporal coverage and interpretive evaluation of relevance, precisely the capacities that HistoRAG's interventions are designed to provide.

A keyword baseline identifies 4,207 articles (4.1\% of corpus) containing at least one computerisation-related term. Within this subset, reader letters constitute only 163 articles (3.9\%), a notable underrepresentation given that reader letters make up 11\% of the full corpus (11,286 articles). This disparity (likely reflecting readers' use of everyday vocabulary over technical terminology) provides a concrete test case for whether RAG-based retrieval can surface sources that lexical filtering misses.

\subsection{System Architecture}
\label{sec:system_architecture}

SPIEGELragged implements the two-phase HistoRAG workflow as a web application with a Python Flask backend and a Next.js React frontend. The backend exposes REST API endpoints for standard semantic search, LLM-assisted search with evaluation, and analysis of selected chunks. The two-phase design (Section~\ref{sec:separated_retrieval}) is realised through separate interface tabs labelled \textit{Heuristik} (source discovery) and \textit{Analyse} (LLM-assisted interpretation) with explicit user-initiated transfer between them.

The retrieval pipeline uses ChromaDB as a vector store with the HNSW (Hierarchical Navigable Small World) algorithm for approximate nearest-neighbour search across approximately 200,000 chunks. Text is chunked using LangChain's RecursiveCharacterTextSplitter at configurable granularities (500--3,000 characters). Embeddings are generated using nomic-embed-text \citep{nussbaum2025NomicEmbedTraining}, an open-source model with 768-dimensional embeddings and an 8,192-token context window, subsequently migrated to nomic-embed-text-v2-moe \citep{nussbaum2025TrainingSparseMixture} for improved multilingual performance on the German-language corpus. All evaluation results reported in Section~\ref{sec:evaluation} use the migrated model. Cosine similarity serves as the distance metric, with the caveats about high-dimensional similarity noted by \citet{you2025SemanticsAngleWhen}.

Temporal windowing (Section~\ref{sec:temporal_windowing}) is implemented as a pre-retrieval partitioning step that divides the search period into configurable windows and allocates retrieval equally across them. LLM-as-judge evaluation (Section~\ref{sec:llm_judge}) supports multiple model backends (including locally hosted models via Ollama as well as commercial APIs from OpenAI, Anthropic, Google, and DeepSeek), enabling the cross-model robustness analysis presented in Section~\ref{sec:evaluation}.

A further component bridges lexical and semantic retrieval through FastText word embeddings \citep{bojanowski2017EnrichingWordVectors} trained on the full corpus. Word embeddings have proven valuable for tracking semantic change in historical texts \citep{hamilton2016DiachronicWordEmbeddings, kutuzov2018DiachronicWordEmbeddings}, and in RAG systems they can serve a dual function that complements the text embeddings used for chunk retrieval. First, they enable \textit{query expansion}. Starting from a user's initial search terms, the system identifies semantically proximate vocabulary as attested within the corpus itself, surfacing period-specific synonyms and related terms that the researcher may not have anticipated. Searching for "Computer", for instance, reveals that "Elektronenhirne" appears with high semantic similarity (0.788) but low corpus frequency (16 occurrences), suggesting rapid obsolescence and guiding the researcher toward vocabulary that would otherwise require manual lexicographic work, or never come into focus at all. Second, corpus-trained word embeddings provide a form of \textit{vocabulary verification} grounded in \textit{DH~1.0} methods. Because the embedding space reflects co-occurrence patterns in the actual source material, the semantic relationships it surfaces can be checked against the corpus, not taken on trust from a general-purpose model. This positions word embeddings as a controlled intermediary between keyword search (which requires knowing the exact terms in advance) and text-embedding-based retrieval (which operates in a high-dimensional space that resists direct inspection). In our implementation, FastText was chosen specifically for its subword information handling, which makes it robust for the morphologically rich German language and capable of generating meaningful vectors even for rare compound words characteristic of the corpus.

All retrieved chunks carry full provenance metadata (article title, publication date, issue number, source URL), ensuring that any passage surfaced by the system remains traceable to its origin, fulfilling the \textit{Quellenbeleg} requirement that grounds our entire architectural~approach.

\section{Evaluation}
\label{sec:evaluation}

We evaluate each of HistoRAG's three interventions empirically, using the SPIEGELragged system and our \textit{Der Spiegel} corpus as testbed. This evaluation demonstrates that each intervention addresses a measurable deficiency in standard RAG; whether the architecture as a whole productively supports actual historical research practice is a separate question that depends on sustained use with concrete research projects (see Section~\ref{sec:limitations}). The evaluation addresses four questions. Does vocabulary drift create measurable temporal bias in standard retrieval? Does LLM-based evaluation capture dimensions of relevance distinct from vector similarity? Are LLM evaluations robust across different models? And does the multi-layer architecture surface sources that neither keyword search nor similarity retrieval alone would identify?

\subsection{Temporal Windowing: Vocabulary Drift as Retrieval Bias}

If vocabulary evolved across three decades of computerisation discourse, then queries using era-specific terminology should retrieve disproportionately from those eras. We test this by constructing three query sets based on \citeauthor{busch2015DiskurslexikologieUndSprachgeschichte}'s periodisation of computerisation vocabulary. These include 1950s terms ("Elektronenhirn", "Denkmaschine", "Rechenmaschine", "Roboter"), 1960s terms ("Automation", "Lochkarte", "Rechenautomat", "IBM", "Datenverarbeitung"), and 1970s terms ("Computer", "EDV", "Gro{\ss}rechner", "elektronische Datenverarbeitung", "Minicomputer"). Each query set was run against the full corpus (1950--1979) using standard cosine similarity retrieval over the sentence embeddings alone, without query expansion or temporal windowing, returning the top 50 most similar chunks.

\begin{figure}[htbp]
  \centering
  \includegraphics[width=0.9\textwidth]{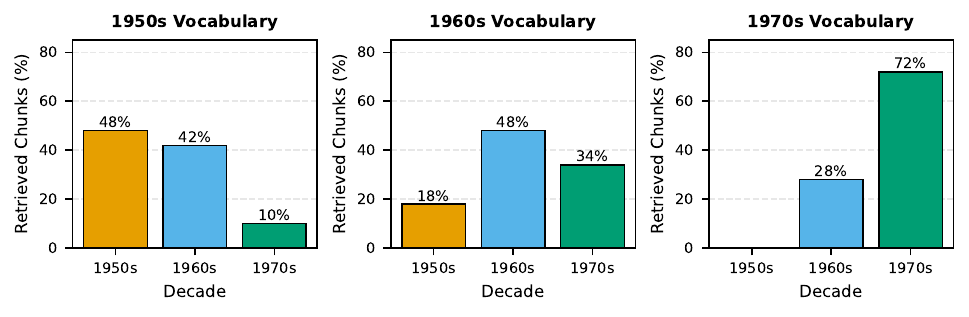}
  \caption{Era-specific vocabulary retrieval distributions. Queries using 1950s, 1960s, and 1970s terminology retrieve disproportionately from their respective eras, with 1970s vocabulary retrieving zero chunks from the 1950s.}
  \label{fig:era_vocabulary}
\end{figure}

The pattern is striking (Figure~\ref{fig:era_vocabulary}). Queries using 1950s vocabulary retrieve 48\% of chunks from the 1950s, double the corpus baseline of 24\%. Queries using 1960s vocabulary peak in the 1960s at 48\%. Most dramatically, queries using 1970s vocabulary retrieve \textbf{zero chunks from the 1950s} and 72\% from the 1970s. A researcher approaching computerisation discourse with contemporary vocabulary would systematically miss the formative period of public discourse formation.

One might assume that combining vocabulary from all periods would solve this problem. We test this by constructing a comprehensive query combining all three vocabulary sets. The comprehensive query without temporal filtering retrieves 28.4\% from the 1950s, 38.4\% from the 1960s, and 33.3\% from the 1970s. At the decade level, this aligns reasonably well with corpus composition (23.6\%, 35.5\%, and 41.0\% respectively); the 1950s and 1960s are slightly over-retrieved and the 1970s slightly under-retrieved relative to their share of the corpus. Corpus composition serves here only as a descriptive reference rather than a target distribution, since the share of articles per period itself reflects the uneven thematic density of the discourse rather than what a temporally balanced sample requires. The five-year breakdown, however, reveals that this aggregate balance hides a finer-grained skew. Only 4.2\% of retrieved chunks come from 1950--1954, against a corpus share of 10.9\% (a retrieval-to-corpus ratio of 0.39), while 29.2\% come from 1965--1969, against a corpus share of 18.7\% (ratio 1.56). The earliest formative period is systematically under-retrieved even when the decade aggregates do not register the bias. The key insight is that query formulation alone cannot solve the temporal bias problem; a researcher inspecting only decade-level retrieval percentages would not see it at all. Constructing a query that did so would itself be highly presuppositional, since it would require anticipating the very temporal distribution the research sets out to discover. Whether using narrow or comprehensive vocabulary, the researcher cannot predict or control how the embedding model will distribute retrieval across time periods. Vocabulary choice becomes an invisible methodological decision with consequences the researcher cannot anticipate.

\begin{figure}[htbp]
  \centering
  \includegraphics[width=0.9\textwidth]{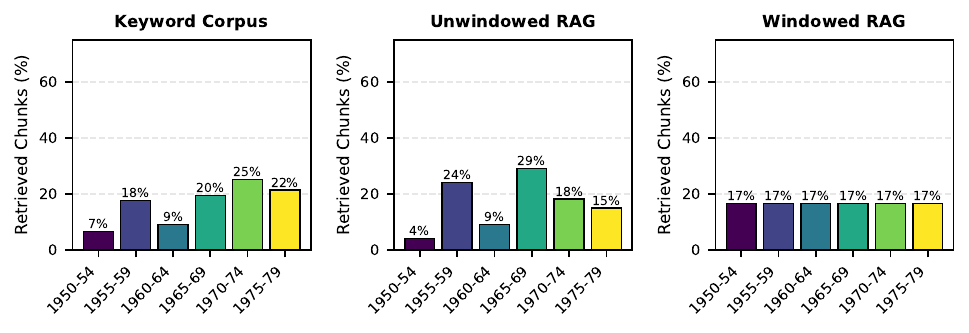}
  \caption{Temporal distributions of keyword filtering, unwindowed RAG retrieval, and windowed RAG retrieval. Only the windowed approach guarantees balanced representation across the research period.}
  \label{fig:temporal_comparison}
\end{figure}

Comparing retrieval with and without temporal windowing makes the consequences concrete (Figure~\ref{fig:temporal_comparison}). Articles surfaced by windowing cluster in the 1950s, while deprioritised articles come primarily from the 1960s, producing a more balanced temporal distribution. The similarity score differential makes the trade-off explicit. Surfaced articles score lower in vector similarity (mean 0.703) than deprioritised articles (mean 0.753). By guaranteeing temporal balance, we accept lower aggregate similarity in exchange for temporal coverage, treating temporal representativeness as a dimension of relevance that similarity metrics alone cannot capture. This distinction between similarity (thematic proximity to the query) and relevance (value for the research question) is central to our architecture and is addressed empirically through LLM-as-judge evaluation in Section~\ref{sec:dual_score}.

In a further test, we compare semantic retrieval against keyword search on computerisation-related terms, the full-text method historians have traditionally used in this domain. This comparison reveals a further pattern. Lexical matching produces an even more pronounced temporal skew than semantic retrieval. Over 61\% of keyword-matched articles come from the 1970s and only 10.7\% from the 1950s. Neither semantic similarity nor lexical matching inherently produces temporal balance; both require explicit methodological intervention to ensure coverage across the research period. Only the windowed approach guarantees the balanced representation that historical research on discourse evolution demands.

\subsection{LLM-as-Judge: Dual Score Analysis}
\label{sec:dual_score}

To evaluate whether LLM-based assessment captures dimensions distinct from vector similarity, we retrieved 120 Leserbriefe (reader letter) chunks across the full search period (1950--1979), using temporal windowing to ensure balanced coverage.\footnote{The retrieval query was a natural language description of the research topic (public hopes and fears regarding computerisation and automation in West Germany), formulated to position the query embedding in the relevant region of semantic space. The LLM evaluation criteria were specified as a detailed German-language rubric scoring sources 1--10 on the degree to which they provide insight into contemporary hopes or fears regarding technological change, with explicit descriptors for each score band.} All 120 retrieved chunks were evaluated, with no manual pre-selection of relevant sources. Each received a vector similarity score from the embedding-based retrieval and an LLM evaluation score from Gemini 2.5 Pro, which assessed it against the historian-defined relevance rubric detailed in the footnote above.

\begin{figure}[htbp]
  \centering
  \includegraphics[width=0.75\textwidth]{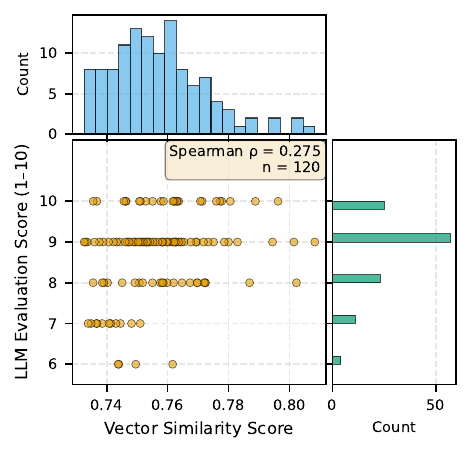}
  \caption{Vector similarity versus LLM evaluation score for 120 reader letter chunks. The weak correlation (Spearman $\rho$ = 0.275) is consistent with the two measures capturing different dimensions of relevance, though compressed score ranges likely attenuate the association (see text).}
  \label{fig:dual_score}
\end{figure}

The correlation between these measures is statistically significant but weak (Pearson $r$ = 0.273, $p$ = 0.003; Spearman $\rho$ = 0.275, $p$ = 0.002). Given the non-normal distribution of both score variables (confirmed by Shapiro-Wilk tests), Spearman's rank-order correlation is the more appropriate measure. Both coefficients indicate that vector similarity and LLM-assessed relevance are statistically related but only weakly so (Figure~\ref{fig:dual_score}). This weak correlation is consistent with a central assertion of our architecture, namely that vector similarity and interpretive relevance capture overlapping but substantially different dimensions. Both score distributions are compressed, however, with LLM scores clustering in the upper range and vector similarities falling within a narrow band, which restricts their range and likely attenuates the correlation. We therefore treat the weak association as supporting this distinction rather than as a precise estimate of it. A source can be semantically proximate to our query while offering limited insight into our research question; conversely, a source discussing anxieties about technological change in vocabulary that differs from our query formulation may prove highly relevant. This divergence should not be read as the LLM evaluation correcting vector similarity, since the LLM score is itself a criterion-based judgment defined by our rubric rather than an independent ground truth. The two are best treated as distinct and contestable signals. The cross-model analysis below speaks to the consistency of these judgments across models, not to their correctness.

The vector similarity scores cluster tightly (range: 0.733--0.808), leaving little room for discrimination, yet within this narrow band, the LLM evaluation identifies meaningful variation in interpretive relevance. The LLM scores cluster heavily in the upper range (87.5\% scoring 8 or above), reflecting the effectiveness of the natural language query formulation. Because the query already encodes a semantically rich description of the research topic, the retrieved chunks are predominantly on-target. The LLM's task is to distinguish \textit{degrees} of relevance among sources already pre-selected for topical proximity, evaluating, for instance, whether a source that \textit{mentions} computerisation fears also \textit{substantively engages} with them.

Examining specific cases illuminates the divergence. The 1956 article "Rationalisierer raus" (Rationalizers Out) received one of the lowest vector similarity scores in our corpus (0.739, below the 25th percentile) yet a high LLM evaluation (9/10). The evaluation explains. "\textit{Der Text thematisiert explizit die Angst vor der Umstellung auf automatischen Betrieb und die Verdr\"angung \"alterer Arbeiter durch Rationalisierung.}" (The text explicitly addresses the fear of the switch to automated operation and the displacement of older workers through rationalization.) The article's vocabulary centres on "Rationalisierung" and workplace conflict (workers shouting "Die Rationalisierer m\"ussen raus!") rather than our query's emphasis on "Computerisierung" and "Automatisierung". Yet its content directly addresses our research question about hopes and fears surrounding technological transformation. This is precisely the kind of source that keyword filtering would miss and that vector similarity undervalues relative to its interpretive relevance.

The reverse pattern is equally instructive. "Unter den schwarzen Kreuzen" (1953) scores above the median for vector similarity (0.762) but receives a comparatively lower LLM evaluation (6/10). Its vocabulary overlaps with our query (\textit{Rationalisierung, technologischer Fortschritt}), positioning it closer in embedding space, yet the LLM recognises that the article treats these themes at a level of generality that provides limited insight into the specific affective dimensions of computerisation discourse. The two scores capture genuinely different aspects of relevance.

\subsection{Cross-Model Robustness}
\label{sec:cross_model}

Although LLM-as-a-judge creates outputs that are visible and contestable, the question of reproducibility remains. If different LLMs produce substantially different evaluations, what does this mean for scholarly practice? We examined this using a focused corpus of 30 reader letters from 1965--1979, evaluated independently by four models: Claude Haiku 4.5, Gemini 2.5 Pro, DeepSeek Reasoner, and Mistral Large (locally hosted via HU Berlin LLM infrastructure).

\begin{figure}[htbp]
  \centering
  \includegraphics[width=0.85\textwidth]{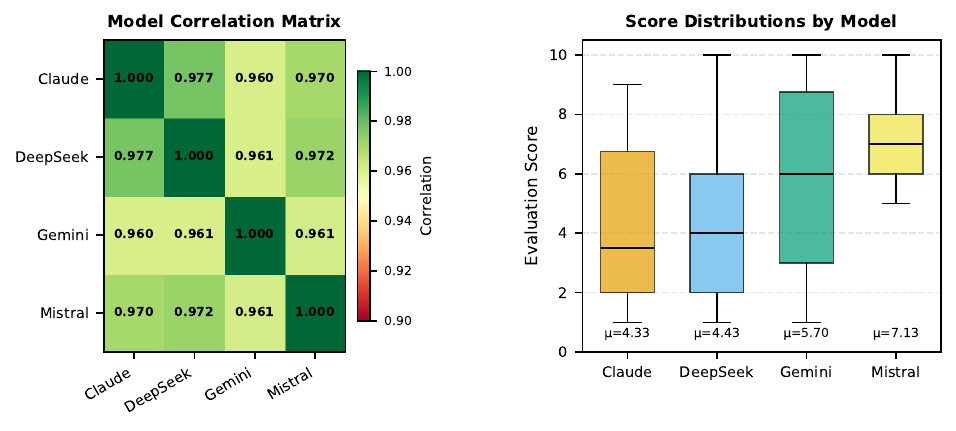}
  \caption{Cross-model evaluation comparison. Left: correlation heatmap showing strong rank-order agreement (r > 0.96) across all model pairs despite different score distributions (right).}
  \label{fig:model_comparison}
\end{figure}

The score distributions reveal substantial calibration differences (Figure~\ref{fig:model_comparison}). Claude and DeepSeek operate as strict evaluators, using the full 1--10 scale with mean scores around 4.3--4.4. Gemini is moderately lenient. Mistral exhibits a notable "floor effect", never assigning scores below 5 even for sources it acknowledges as thematically irrelevant. Its evaluation texts for low-scoring sources often contain phrases like "\textit{obwohl kein direkter Bezug}" (although no direct connection) while still assigning 5/10.

Yet beneath this calibration variation, ranking agreement is strong. All model pairs show correlations above 0.96 on this sample, indicating substantial agreement on which sources are more or less relevant relative to each other. The disagreement lies in absolute calibration, not relative ranking. Claude and DeepSeek agree within 1 point on 100\% of chunks; Claude and Mistral agree within 1 point on only 27\%, yet they rank sources in nearly identical order ($r$ = 0.970). The models share an underlying assessment of relevance but differ in how they map that assessment onto numerical scales. We note that this agreement is partly facilitated by the small sample size ($n$=30), which includes clear-cut cases at both ends of the relevance spectrum. With larger corpora containing more borderline cases, we would expect greater divergence in edge-case judgments, and these results should be treated as indicative rather than definitive.

This finding has important methodological implications. These results indicate that rankings are largely portable across models, while absolute score thresholds are not. A filtering rule like "retain all sources scoring 7 or above" would yield dramatically different corpora depending on which model performed the evaluation. But the question "which sources are most relevant for our research question?" receives consistent answers regardless of model choice. To circumvent the differing scorings, we suggest working with percentiles rather than absolute thresholds, as these remain constant across models.

The disagreement cases are instructive. A 1971 letter about Japanese economic competition lacks direct relevance to computerisation. Claude, Gemini, and DeepSeek express this as scores of 1--2; Mistral gives 5/10 despite acknowledging "\textit{...fehlt jeglicher Bezug zu Technologie.}" The models agree on substance but differ on how generously to score sources outside the research focus. For the historian, the evaluation texts are more informative than the scores themselves, as all four models explain \textit{why} the source is peripheral to the inquiry. A more comprehensive cross-model evaluation with larger samples and borderline cases is planned as part of ongoing work with the HistoRAG architecture.

\subsection{Retrieval Layers in Practice: What the Pipeline Surfaces}

To illustrate how the multi-layer architecture functions in practice, we examine what the retrieval+evaluation pipeline delivers for Leserbriefe on our hopes-and-fears query, and how the results distribute against keyword-corpus membership. This is a within-pipeline analysis, not a head-to-head comparison of retrieval methods: such a comparison would require chunk-level evaluation of both corpora under matched selection mechanisms, which we defer to forthcoming work (see caveats below). Section~\ref{sec:corpus} established that the keyword corpus contains 163 reader letters (3.9\% of 4,207 articles), a significant underrepresentation relative to the 11\% proportion in the general corpus.

In a separate retrieval run using our hopes-and-fears query with temporal windowing and LLM-as-judge evaluation, we retrieved 150 Leserbriefe chunks across the full search period (50 per decade through temporal windowing). Each chunk was evaluated using our LLM criteria. Because RAG retrieves chunks while the keyword corpus is defined at article level, we aggregate to the article level by taking each article's maximum chunk score. This is an imperfect bridge between two different units of work and is one of the reasons a proper cross-method comparison is deferred (see below); the purpose here is to characterise what the pipeline's output looks like, not to compare methods at matched granularity.

The overlap between the two retrieval sets is smaller than expected. Only 21 articles (12.9\% of the keyword corpus) appear in both. The two methods capture largely disjoint article sets. Within the RAG-retrieved set, the next question is how the LLM evaluation distributes relevance across articles that are also in the keyword corpus versus articles outside it, since this distribution clarifies which sources the semantic layer is adding to the pipeline's reach.

\begin{figure}[htbp]
  \centering
  \includegraphics[width=0.75\textwidth]{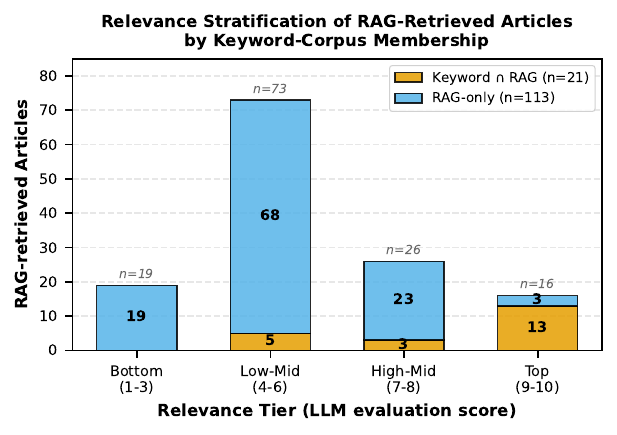}
  \caption{Relevance stratification of RAG-retrieved articles (n=134) by keyword-corpus membership. Keyword-overlap articles (n=21) cluster in the top tier; RAG-only articles (n=113) span the full range, with all 19 bottom-quartile articles being RAG-only. The figure characterises what the pipeline surfaces and is not a cross-method recall comparison (see caveats in text).}
  \label{fig:quality_stratification}
\end{figure}

The two distributions have very different shapes (Figure~\ref{fig:quality_stratification}). Of the 21 articles in the keyword-RAG overlap, 16 (76\%) score in the top quartile (scores $\geq$ 7, the threshold at which the evaluation criteria indicate clear relevance to the research question as opposed to tangential or partial engagement), and 13 of those (62\% of all overlap articles) score 9-10. The keyword-overlap distribution is heavily top-loaded. The 113 RAG-only articles cluster in the middle of the relevance scale instead, with only 26 (23\%) reaching the top quartile and just 3 reaching 9-10. At the other end, all 19 bottom-quartile articles (scores below 4) are RAG-only, confirming that semantic retrieval expands coverage but introduces noise requiring the LLM evaluation filter.

Two observations follow. First, when keyword search hits a Leserbrief that semantic retrieval also surfaces, the article is very likely to address the research question directly. This is a within-overlap observation (the n=21 intersection, not the full 163-article keyword corpus, whose remaining 142 articles were not LLM-evaluated under this protocol). A full cross-method comparison would require evaluating the entire keyword corpus, a dedicated study in its own right and beyond the architectural focus of this paper. This observation should not be read as a claim about keyword search's precision in general. Second, of the 134 articles the pipeline surfaced, 26 high-relevance articles are outside the keyword corpus entirely. They use vocabulary like "Rationalisierung" or "Arbeitslosigkeit" rather than explicit computerisation terminology, and are therefore unreachable by keyword filtering under the scheme used here. This demonstrates that the semantic layer extends the pipeline's reach into vernacular discourse, but it does not establish that semantic retrieval finds more high-relevance sources than keyword search in absolute terms.

One might ask whether those 26 sources could be captured simply by extending the keyword list. Many of the RAG-retrieved letters use terms like "Rationalisierung", "technische Revolution", "Arbeitslosigkeit", or "Elektronik", vocabulary a more expansive keyword scheme would match. The architectural question, however, is what expanding the keyword list entails. Adding terms like "Arbeitslosigkeit", "technisch", "Industrie", or "Fortschritt" across 102,189 articles over thirty years would match thousands of additional articles with no connection to computerisation discourse. The keyword approach faces a precision-recall trade-off that worsens progressively as vocabulary broadens. Semantic retrieval combined with LLM-as-judge evaluation navigates this trade-off differently, casting a broader net through conceptual proximity and filtering by interpretive relevance to the specific research question. Whether semantic retrieval outperforms keyword extension for a given research task is an empirical question that depends on the corpus, the query, and the scoring protocol, and is precisely the kind of comparison we defer to dedicated future work.

Two points follow for the architecture. First, because keyword search and semantic retrieval surface largely disjoint article sets (21 overlap out of 276 when combined), a pipeline that uses both highlights candidates neither would surface alone. Our architecture accordingly incorporates keyword search (with corpus-based vocabulary expansion) as a first retrieval layer alongside semantic retrieval. Second, the noise that semantic retrieval introduces in its share of the candidate pool (87 of 113 RAG-only articles fall below the relevance threshold) is what makes LLM-as-judge evaluation essential rather than optional for any pipeline that casts a broader net. This second finding is the one we regard as strongest, because it does not depend on cross-method comparison: regardless of how candidates enter the pipeline, a post-retrieval evaluation layer that applies explicit, contestable criteria is necessary to separate interpretively relevant sources from thematically adjacent noise.

For the scale of research this architecture is designed to support, the system's primary contribution to the research process is \textbf{prioritisation}, not discovery. The keyword corpus alone contains 163 Leserbriefe, and the semantic layer adds 113 more candidates outside that corpus. Close reading the full union (276 articles) for a single query is feasible but costly. The LLM evaluation layer triages the pool, producing a ranked working corpus with a written justification for each score, and the historian retains authority to override any judgment in either direction. The system augments scholarly judgment without replacing it.

Several caveats bound the scope of this analysis. First, the unit of comparison differs between the two methods. Keyword search selects whole articles (presence of a term anywhere in the body); RAG retrieves chunks scored against a natural-language query. Aggregating RAG output to articles via max chunk score, as we do here, bridges the two units but does not equalise them: the LLM's article-level score on a keyword article is not directly comparable to the LLM's chunk-level score on a RAG-retrieved passage, because the text the evaluator sees is different in length, selection mechanism, and contextual embedding. A proper cross-method comparison requires chunk-level scoring with matched chunking and selection protocols, including keyword matching applied at the chunk rather than article level, which we regard as the subject of dedicated methodology work rather than an architecture paper. Second, exploratory LLM scoring of the 142 keyword-matched Leserbriefe that semantic retrieval did not surface suggests those articles are also predominantly high-relevance. This means the counts reported here cannot support a recall comparison between methods in either direction, and we do not make one. Third, the balance between keyword and semantic retrieval is topic-dependent. Our computerisation query benefits from relatively clear, distinctive terminology, meaning the keyword baseline performs comparatively well. For research topics with more ambiguous vocabulary, extensive synonymy, or concepts expressed only implicitly, the precision-recall problem of keyword extension would be more acute and the advantage of semantic retrieval correspondingly more pronounced. None of this changes the architectural argument of the paper, which is that a pipeline combining both retrieval layers under a post-retrieval LLM evaluation filter is more robust than any single layer, and that the evaluation layer itself is essential regardless of which retrieval method sources candidates.

\section{\textit{Zwischentexte}: LLM Outputs as Interpretive Proposals}

The preceding sections established an architecture that transforms RAG from a seamless question-answering pipeline into a structured research process, addressing how sources are found, balanced, and evaluated. But one question remains open, namely what happens when the system generates text. RAG systems typically pass retrieved context to an LLM and produce text. What changes with our system is not the technical mechanism but the kind of question we pose and, consequently, the role the output plays in historical knowledge production.

In standard RAG, the generation step is oriented toward factual question-answering. A user asks a question, and the system produces an answer grounded in retrieved passages. Our architecture repurposes this generation step. Rather than asking what the sources say about a specific fact, we ask how they speak about a theme over time, prompting the model to identify thematic connections, trace argumentative patterns across decades, and propose interpretive structures. These outputs are what we call \textit{Zwischentexte} (intermediate texts that serve neither as answers nor as finished scholarship but as interpretive scaffolding). The term captures their epistemic status. They lie between retrieved sources and historical argument, offering first proposals for interpretation that the historian can verify, contest, and develop. They are tools embedded within the research process, designed to surface patterns and connections at corpus scale, not to deliver conclusions.

This reframing points toward what is perhaps the central question for LLMs in digital humanities more broadly. The computational methods of \textit{DH~1.0} identify formal patterns on the basis of predefined rules and features. Their outputs are quantitative indicators (frequency distributions, similarity scores, network graphs) that require the historian to supply interpretation. Zwischentexte, as products of \textit{DH~2.0}, operate differently. They deliver interpretive suggestions that emerge from the complex, opaque processes of language modelling. Unlike a word frequency list, a Zwischentext proposes meaning. It claims that a term functions as a "semantic battleground", that anxiety "migrates" through the class structure, that 1964 marks a discursive rupture. These are not formal patterns influencing interpretation but interpretive proposals requiring validation. Drawing on Droysen's distinction between \textit{Erkl\"aren} (explaining) and \textit{Verstehen} (understanding) \citep{maclean1982JohannGustavDroysen}, we might say that LLMs can provide explanations (identifying patterns, synthesising connections across texts, articulating relationships) but the task of \textit{Verstehen}, a deeper historical understanding that defines history as a hermeneutic science \citep{gadamerWahrheitUndMethode}, remains with the historian.

To test this framework, we generated Zwischentexte from a curated corpus of 65 Leserbriefe (drawn from the 150 chunks retrieved in Section~\ref{sec:evaluation})\footnote{Several reader-letter selections appear across the evaluation, distinguished by purpose and not to be read as a single evolving sample. They are the 120 chunks used for the dual-score correlation (Section~\ref{sec:dual_score}), the 150 chunks (50 per decade) used for the retrieval-layer analysis, the 65-letter high-relevance subset of the latter used here for Zwischentext generation, and the 163-letter keyword corpus (Section~\ref{sec:corpus}).} using four interpretive queries addressing our metaquestions: concrete hopes expressed by readers, concrete fears about automation, turning points in sentiment across decades, and vernacular uses of the term "Rationalisierung". Comprehensive synthesis texts integrating findings across all four queries were generated by frontier models (Opus 4.5, Sonnet 4.5, GPT-5.1). Gemini was not included, since in preliminary use it did not show the same facility with close textual analysis as the other models. The curated corpus itself embodied the historian-in-the-loop principle. Chunks scoring 7--10 on the LLM evaluation were retained automatically; those scoring 0--2 were excluded, and the crucial middle range (3--6) was reviewed individually, with the historian examining each evaluation justification and making a judgment call about inclusion.

Validation through close reading of the resulting Zwischentexte confirmed that LLM-generated interpretations can be both surprisingly accurate and subtly misleading. Consider the claim, appearing across multiple synthesis texts, that 1964 marks a "Stimmungsumschwung" (mood shift). The Zwischentext cites a letter describing how SPIEGEL's coverage "erzeugt praktisch eine Panikstimmung." We located this source through the provided URL and verified the full passage. Hans Benzinger of N\"urnberg criticises an internal contradiction within SPIEGEL's own coverage (the editorial creates panic while an interview downplays it). The citation is accurate, and the full context enriches the interpretation, suggesting that media framing itself was contested terrain, not merely a reflection of pre-existing sentiment. We also verified a 1955 Thorneycroft quotation about "menschenleere Betriebe" that opens several synthesis texts; the narrative detail of the raised glass is not invented but comes directly from the source. This verification matters because compelling narrative details are precisely where hallucination risk is highest.

Where historian judgment diverges, the divergence is itself instructive. The Zwischentexte consistently interpret "Rationalisierung" through a contemporary lens, as if it straightforwardly meant "automation" or "technological unemployment". But for 1950s readers, the term carried specific lived-historical resonances from the Weimar-era rationalisation movement \citep{shearer1995TalkingEfficiencyPolitics} that the LLM cannot recover from the retrieved sources alone. The quality of LLM-assisted analysis depends not on the machine alone but on the historian's expertise and ability to supply contextual knowledge through targeted prompting. This underscores that Zwischentexte are starting points for inquiry, not endpoints. Their strongest use is hypothesis generation. A claim that 1964 represents an earlier anxiety rupture than previously recognised is not proven by the Zwischentext; it is proposed. The historian can then test this hypothesis through targeted close reading, contextual research, and engagement with existing scholarship. Zwischentexte that fail this test are expected outcomes of a process designed to produce contestable proposals. In this case, the proposals also engage productively with existing historiography. \citeauthor{schuhmannTraumVomPerfekten}'s \citeyearpar{schuhmannTraumVomPerfekten} foundational study of computerisation discourse traces a narrative arc from early euphoria to late-1970s anxiety, identifying the 1978 SPIEGEL cover "Uns steht eine Katastrophe bevor" \citep{1978UnsStehtKatastrophe} as emblematic. Our Zwischentexte confirm this general arc but suggest that the discursive rupture occurred earlier than canonical accounts recognise, with multiple reader letters responding to a 1964 automation cover story in language of existential anxiety. Whether this earlier dating withstands further scrutiny is precisely the kind of question the Zwischentexte are designed to generate.

These findings connect to a broader question about the status of LLM-generated text in scholarly practice. Recent work has shown formally that hallucination is an innate limitation of large language models, not a flaw to be engineered away \citep{xu2025HallucinationInevitableInnate}, which strengthens the case for treating LLM outputs as proposals requiring validation, not as reliable findings. As LLMs become capable of producing fluent, plausible academic prose, the risk is not that scholars will be replaced but that the distinction between generated text and warranted argument will erode. These Zwischentexte make this distinction explicit by design. They are labelled as machine-generated interpretive proposals, they carry source citations that can be verified, and their evaluation criteria are transparent. This contrasts with scenarios where LLM outputs are silently incorporated into scholarly writing without acknowledgment of their provenance or epistemic status. The concept of Zwischentexte thus offers one model for how humanities disciplines might integrate LLM-generated text responsibly, as visible, contestable intermediate products within a documented research process.

At a base level, Zwischentexte are signs, not meaning \citep{hiltmann2024HermeneutikZeitenKIa}. This is no contradiction of the shift from \textit{DH~1.0} to \textit{DH~2.0} described in the introduction. What changes with LLMs is the level at which our interaction with computation operates, approximating meaning rather than counting sequences of signs. What does not change is the hermeneutic status of the text that results, generated or otherwise, whose meaning is never contained in the artefact itself. The LLM produces tokens that represent words, patterns, connections. Out of these signs, the historian produces meaning through interpretation grounded in context, method, and scholarly judgment. The division of labour this implies is significant but not unprecedented. Historians have always relied on tools that pre-process sources, from archivists who organise collections to editors who transcribe manuscripts and indexers who create finding aids. Each intermediary makes interpretive decisions that shape what historians can discover. The difference with LLMs is scale and opacity. They process vastly more text but through mechanisms we cannot fully inspect. The architecture we have described attempts to manage this trade-off by preserving transparency where it matters most, in the criteria for relevance, the justifications for evaluation, and the sources behind any claim.

\section{Discussion}

The preceding sections have demonstrated that each of HistoRAG's three interventions addresses a measurable deficiency in standard RAG and that their combination produces a retrieval architecture qualitatively different from generic pipelines. We now turn to what these findings mean for the broader question of how computational systems should be designed for interpretive scholarship.

\subsection{What Generalises}

Our three interventions are not specific to \textit{Der Spiegel} or to computerisation discourse. Separated retrieval and generation implements a distinction (between finding sources and interpreting them) that applies to any research process where corpus construction is itself a scholarly act. Temporal windowing addresses a structural property of embedding-based retrieval: vocabulary drift creates temporal bias whenever a corpus spans periods of linguistic change. Any historical RAG application working across decades will face this problem, whether the subject is computerisation, colonialism, or climate discourse. LLM-as-judge evaluation, likewise, generalises wherever "relevance" is not a fixed property but depends on the research question being asked, which is to say, wherever scholarship is interpretive rather than factual.

The principles extend beyond history. Legal scholarship navigating case law across jurisdictions and decades, literary analysis tracing stylistic evolution through a corpus, and qualitative social science synthesising interview transcripts all face analogous tensions between computational retrieval defaults and disciplinary requirements. The specific criteria would differ (a legal scholar's evaluation rubric foregrounds precedential weight, not affective framing), but the architectural pattern (broad retrieval, temporal or contextual balancing, criteria-based interpretive filtering) transfers.

\subsection{On Agentic Systems}
\label{sec:agentic}

A word on what we have deliberately \textit{not} done. Contemporary AI development emphasises "agentic" systems, architectures where LLMs autonomously plan, execute, and iterate through complex tasks with minimal human intervention. We have resisted this direction. Not because agentic systems lack capability but because autonomy and transparency exist in tension. Each decision an agent makes (to search this rather than that, to prioritise these results, to frame the synthesis in this way) is a decision the historian does not see, cannot evaluate, cannot contest. The efficiency gained comes at the cost of epistemic agency transferred to processes that operate opaquely.

This does not mean agentic approaches have no place in historical research. For well-defined tasks with clear correctness criteria (formatting references, identifying duplicate documents, transcribing standardised forms) autonomous processing may be appropriate. But for the interpretive core of historical work (deciding which sources matter, assessing what they mean, constructing arguments about the past) the historian must remain in the loop, not as a rubber stamp for machine decisions but as the locus of scholarly accountability.

There is a further point. The architectural principles we have introduced encode methodological commitments that produce better research independently of who or what executes them. A system that enforces temporal balance retrieves more representative sources than one that does not, whether a historian reviews the results or an agent processes them further. Evaluation criteria that foreground interpretive relevance over vector similarity yield better corpora regardless of the downstream workflow. Getting the architecture right now is therefore not only about preserving space for human judgment in the present but about encoding disciplinary values into computational infrastructure that may outlast any particular division of labour between historian and machine.

\subsection{Limitations}
\label{sec:limitations}

Several limitations constrain the scope of our claims. Computational cost is significant. Evaluating retrieved chunks across temporal windows involves API expenses and processing time that raise questions of sustainability for any workflow routing large volumes of text through LLM APIs. The calibration variation across models (Section~\ref{sec:cross_model}) means that absolute score thresholds are model-dependent. Our relevance stratification using a threshold of 7/10 would yield different counts with a different model or threshold. Because the value 7 is anchored to the rubric's band for clear relevance, it remains an interpretable cutoff within a single model's scores, as in the stratification reported here. Percentiles, which apply equally within a single model, become the more robust choice when comparing across models, where absolute scores are not directly comparable (Section~\ref{sec:cross_model}). One direction for future work is to anchor the evaluation rubric more explicitly, fixing a baseline score as the midpoint and instructing the model how far to adjust upward or downward from it, in the manner of anchored review scales. Such anchoring might reduce the calibration divergence we observe across models.

Query formulation proved to be a consequential methodological variable. Our shift from keyword enumerations to natural language queries aligned with the embedding model's training regime produced substantially higher similarity scores, confirming that query design is itself an interpretive act that shapes what the archive makes visible. Systematic comparison of query formulation strategies across different corpora and research questions remains future work, as does controlled comparison isolating the contribution of each pipeline component under matched conditions.

A deeper concern relates to the alignment training that shapes commercial LLMs. When an aligned model evaluates a 1950s text about "Rationalisierung", its assessment is filtered through contemporary interpretive frameworks embedded during training. More broadly, the model's responses are themselves shaped and altered by the human feedback used to align it. From a hermeneutic perspective, interpretation always requires a \textit{Vorverst\"andnis} (pre-understanding), and this is not a flaw but a condition of understanding. Yet with commercial models, this pre-understanding is neither transparent nor controllable. This is an argument for exploring open-weight or domain-specific models where interpretive assumptions could be made explicit and scholarly contestable.

Our empirical evaluation focuses on Leserbriefe (reader letters) as a test case, chosen because their vernacular vocabulary makes them harder to find through keyword search and thus demonstrates the architecture's potential more fully. However, reader letters are a specific genre (shorter, more affective, more likely to use everyday vocabulary), and the findings may not transfer straightforwardly to longer analytical articles or editorial coverage. Evaluating the architecture across article types remains future work.

The architecture has not yet been evaluated through user studies with practising historians. While we have demonstrated its methodological rationale and empirical consequences, whether the two-phase workflow productively supports actual research practice (and does not merely impose friction) is a question that can only be answered through sustained use. Case studies applying the HistoRAG architecture to the \textit{Der Spiegel} corpus and additional datasets are currently underway.

Finally, our evaluation is based on a single corpus, a single language, and a single research domain. While we have argued that the architectural principles generalise, demonstrating this across corpora, languages, and disciplines remains an important next step.

\section{Conclusion}

We began with a question. How do we preserve historical scholarship's commitment to source sovereignty, interpretive authority, and transparent argumentation when working with AI systems that can "read" thousands of documents and generate plausible-sounding analyses?

Our answer has been architectural, not prohibitory. We do not argue against using LLMs but for \textit{designing} their use around disciplinary values. Separated retrieval and generation restores space for source inspection and reflective corpus construction, allowing informed decisions about what enters the analysis and why. Temporal windowing ensures balanced representation across time periods, countering the vocabulary alignment bias that similarity-based retrieval embeds by default. LLM-as-judge evaluation makes relevance judgments transparent and contestable, producing argued assessments instead of opaque scores. And Zwischentexte (the outputs of this architecture) are treated as interpretive proposals requiring validation and contextualisation that only the historian can provide.

The evaluation offers a first empirical validation that these principled design choices have measurable consequences. Era-specific vocabulary retrieves zero chunks from the 1950s when using 1970s terminology. Vector similarity and LLM-assessed relevance correlate only weakly (Spearman $\rho$ = 0.275). And in our Leserbriefe test case, keyword-based and semantic retrieval surface largely disjoint source pools, with the LLM evaluation layer making relevance judgments transparent across candidates from either method. These results are preliminary, drawn from a single corpus and genre, but they indicate that the architecture measurably changes what the system surfaces to the historian, a precondition for what they can ultimately know.

Standard RAG is optimised for question-answering rather than discourse analysis, for factual retrieval rather than interpretive synthesis, and for seamless user experience rather than epistemic transparency. Historical methodology demands different priorities. Implementing those priorities requires technical choices that embed disciplinary values into system architecture. We offer HistoRAG as one contribution to the emerging practices around AI in historical research, not as the final word but as an opening in a conversation that will define how interpretive disciplines engage with computational methods.

The tools should not determine the questions. The questions should determine the tools. And the historian should determine both.

\section*{Code and Data Availability}
\label{sec:availability}

Both codebases are released under the MIT licence. The reusable framework is distributed as the Python package \verb|historag| at \url{https://scm.cms.hu-berlin.de/digital-history/digital-history-forschung/histo_rag/historag} (this paper describes release \verb|v0.1.0|), and the \textit{Der Spiegel}-specific consuming application SPIEGELragged at \url{https://scm.cms.hu-berlin.de/digital-history/digital-history-forschung/histo_rag/spiegelragged} (release \verb|v1.0-paper|). Archived snapshots of both tagged releases are deposited on Zenodo, at \url{https://doi.org/10.5281/zenodo.20558762} for \verb|historag| and \url{https://doi.org/10.5281/zenodo.20558950} for SPIEGELragged. Both repositories remain under active development beyond the versions described here; later versions add further embedding providers, with Qwen3-Embedding-8B now the default, and are migrating the vector store from ChromaDB to Qdrant. The evaluations in Section~\ref{sec:evaluation} were run with nomic-embed-text-v2-moe embeddings in ChromaDB. The \textit{Der Spiegel} corpus itself, including the source texts and the vector store, is not redistributed, as the archive is held under licence by Spiegel-Verlag; researchers wishing to reproduce our experiments must obtain the corpus from the publisher and run the tagged release's ingestion script (\path{spiegelragged/backend/scripts/ingest_qwen.py}) with that embedding configuration against their own copy. The architectural and methodological contributions remain unrestricted under MIT; the empirical findings reported here are reproducible \textit{modulo} corpus access.

\bibliographystyle{acl_natbib}
\bibliography{references}

\end{document}